\documentclass[twoside, preprint]{article}
\usepackage{jmlr2e} 

\usepackage{amsmath}
\usepackage{graphicx}
\usepackage{booktabs}
\usepackage{hyperref} 
\usepackage{natbib}

\begin{document}

\title{ReCA: A Parametric ReLU Composite Activation Function}

\author{\name John Chidiac \email john.chidiac@lau.edu \\
	\addr Department of Computer Science and Mathematics\\
	Lebanese American University, Jbeil, Lebanon
	\AND
	\name Danielle Azar \email danielle.azar@lau.edu.lb \\
	\addr Department of Computer Science and Mathematics\\
Lebanese American University, Jbeil, Lebanon}

\editor{}

\ShortHeadings{ReCA, a parametric ReLU composite activation function}{Chidiac and Azar}
\firstpageno{1}

\maketitle

\begin{abstract}
	Activation functions have been shown to affect the performance of deep neural networks significantly. While the Rectified Linear Unit (ReLU) remains the dominant choice in practice, the optimal activation function for deep neural networks remains an open research question. 
	In this paper, we propose a novel parametric activation function, ReCA, based on ReLU, which has been shown to outperform all baselines on state-of-the-art datasets using different complex neural network architectures.
\end{abstract}

\section{Introduction}
\label{sec:headings}

Activation functions are mathematical transformations applied to the input signal of neurons
within neural networks. They are essential for introducing non-linearity and allowing the model to learn complex non-linear 
relationships in the data \citep{book}. Without such functions, neural networks would be restricted to linear transformations, limiting their capacity 
to modeling simple relationships only.

Traditional activation functions have a fixed mathematical form involving the input variable (for example, $\phi(x)=sign(x)$), which maintains the same behavior throughout the network. Parametric activation functions
build upon them by introducing parameters that can be learned through backpropagation. These parameters are applied per neuron, which typically grants higher performance at the cost of time, or per channel\footnote{A channel refers to a single feature map or layer of data, such as the color components (e.g., RGB) in an image, that is processed independently during convolution operations.},
which is significantly computationally cheaper while maintaining good performance. Moreover, parametric functions provide additional flexibility to the network and enable specialized per-task learning; however, these parameters 
may lead to increased computational costs compared to simple traditional functions.

We propose a novel parametric activation function, ReCA, which consistently demonstrates significant performance gains on state-of-the-art benchmarks, but at the cost of increased training time. We define ReCA as follows:

\begin{equation}
	f(x)=\alpha \text{ReLU}(x) \left(\left(\frac{1+\tanh(x)}{2}\right)^\beta+\sigma(x)^\delta\right)
	\label{eq:reca}
\end{equation}

Where $\alpha\in(0,+\infty)$, $\beta,\delta\in[0,+\infty)$. This function arose to combine the most beneficial properties of sigmoid, tanh, and ReLU.
The $\tanh$ and $\sigma$ terms provide
fine-tuning over the smoothness of the curve, and the increase of their parameters emphasizes this smoothness even more. Initially,
we set $\alpha=0.5$ and $\beta=\delta\approx0$ such that the function is exactly ReLU, and we allow the parameters to change through backpropagation. The derivative of ReCA is

\begin{equation}
	f'(x)=\begin{cases}
		\alpha f(x)+ \alpha x\left(\frac{\beta \text{sech}^2(x)(\tanh(x)+1)^{b-1}}{2^b}+ \delta e^{-x}\sigma(x)^{d+1}\right) &\text{if } x > 0
		\\
		0 & \text{otherwise}
	\end{cases}
	\label{eq:dreca}
\end{equation}

\begin{figure}[!ht]
	\centering
	\begin{minipage}[b]{0.49\linewidth}
		\centering
		\includegraphics[width=0.98\linewidth]{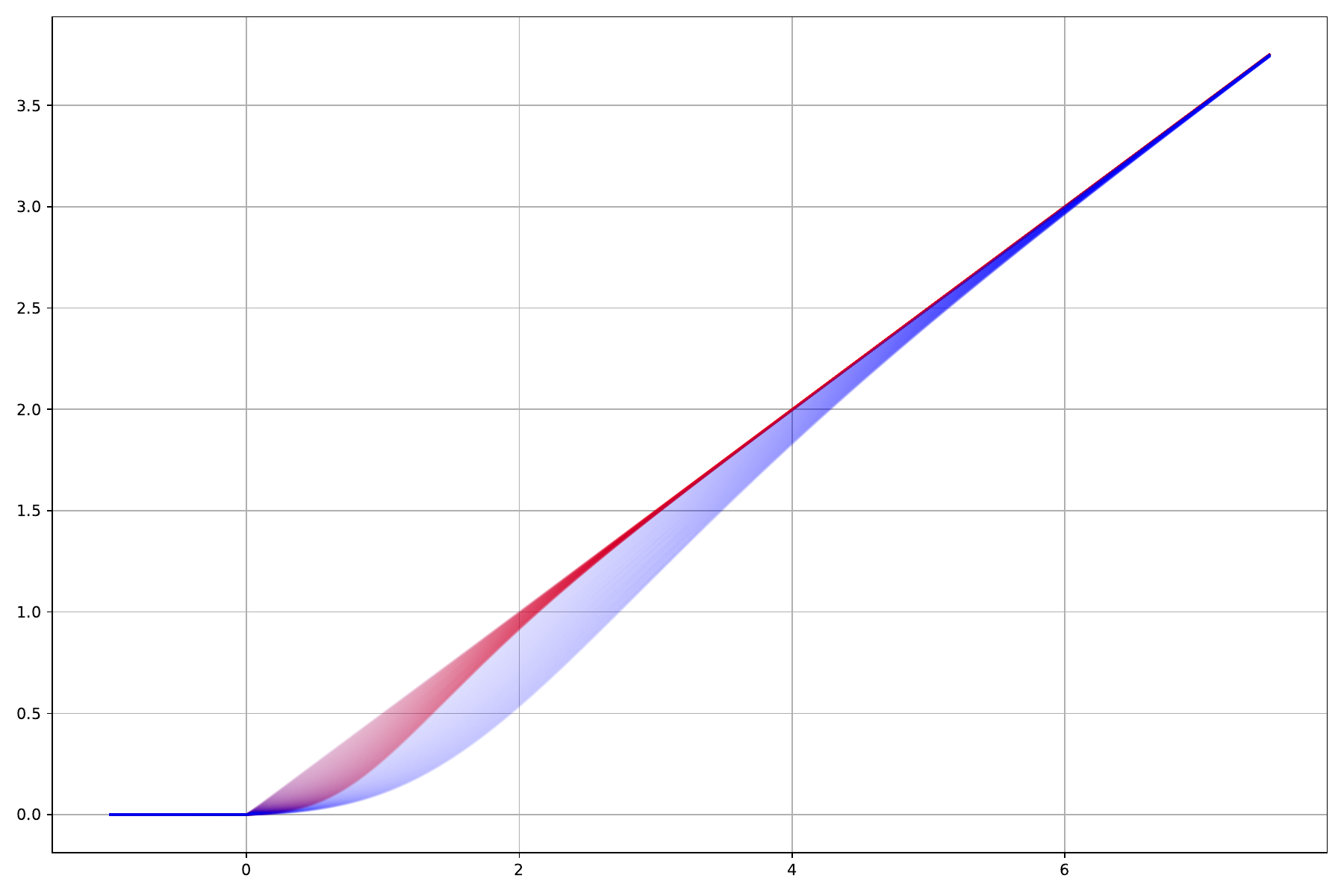}
		\caption{Plot of ReCA showing the effect of $\tanh$ (red) with varying $\beta$ ($\delta=0$) and sigmoid ($\sigma$, blue) with varying $\delta$ ($\beta=0$), for $\beta, \delta \in [0,5]$.}
		\label{fig:components}
	\end{minipage}
	\hfill
	\begin{minipage}[b]{0.49\linewidth}
		\centering
		\includegraphics[width=1\linewidth]{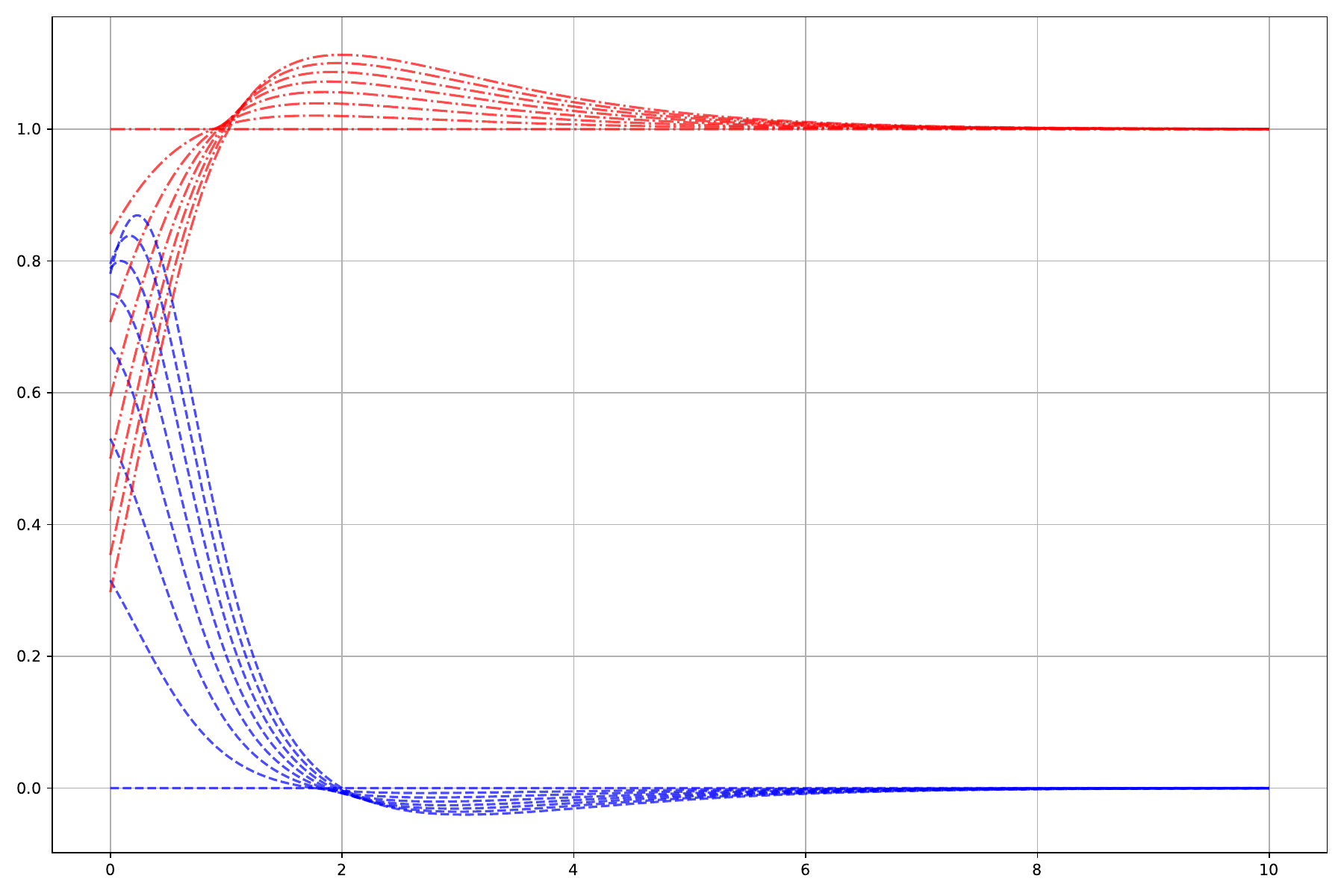}
		\caption{First derivative (red) and second derivative (blue) of the ReCA function, plotted for $\beta = \delta \in (0, 2)$ with a step size of 0.25.}
		\label{fig:derivatives}
	\end{minipage}
\end{figure}

ReCA is monotonic and increasing (as long as $\alpha>0$ and $\beta,\delta\geq0$). We can consider ReCA as a function that controls the degree of linearity of ReLU as shown in Figure \ref{fig:components}.

\section{Related Work}

In recent years, multiple activation functions have been proposed, each offering distinct characteristics that make them suitable
for specific applications. The sigmoid activation function, denoted as $\sigma(x)=\frac{1}{1+e^{-x}}$ introduced by \cite{McCulloch1943} is known for its ability to map 
inputs to a range between 0 and 1, making it suitable for binary classification tasks; however, the function suffers from vanishing gradients
when the inputs are too small or too large which effectively prevents weight updates during backpropagation making it unsuitable for deep neural
networks. The hyperbolic tangent function, $\tanh(x)$, maps values to the range $(-1, 1)$ and leads to better convergence due to the zero-centered output
which helps in stabilizing training and avoiding the saturation of neurons, but the function still suffers from vanishing gradients at extreme input values.
The rectified linear unit activation function, $\text{ReLU}(x) = \max(0, x)$ popularized by \cite{Nair2010RectifiedLU}, has become a default and established activation function for many neural network architectures due
to its speed, efficiency, and effectiveness in mitigating the vanishing gradient problem. Unlike $\sigma$ and $\tanh$, ReLU remains unbounded
for positive values. ReLU creates sparse representations by outputting zero for all negative inputs, which leads to more efficient memory usage and reduced computation time. This also encourages the neural network to learn more 
efficient representations of the data by encouraging the most relevant neurons to be active, which acts as a form of regularization. 
However, this also causes the dying neuron problem, where neurons become permanently inactive by consistently
outputting zero for all inputs, making them unable to learn or recover, since their gradient becomes zero.

Two activation functions emerged to address the dying ReLU problem, the first of which is Leaky ReLU, defined in equation \ref{eq:lrelu}. The $\alpha$ parameter is a small
positive constant, often set to 0.01. This ensures negative inputs still have a small gradient, which reduces the chance of neurons dying. Parametric ReLU, PReLU, expands 
on LReLU by making the $\alpha$ term learnable through backpropagation, which allows the network to adaptively learn the slope for negative inputs, which 
may lead to improved performance \citep{he2015delvingdeeprectifierssurpassing}.

\begin{equation}
	\text{LReLU}(x) = \text{PReLU}(x) =	\begin{cases}
		x & x > 0 \\
		\alpha x & \text{otherwise}
	\end{cases}
	\label{eq:lrelu}
\end{equation}

The swish activation function, defined as $swish(x)=x\cdot\sigma(x)$ provides a smooth, non-monotonic behavior that has been empirically shown to outperform ReLU on a variety
of deep learning tasks \citep{ramachandran2017searchingactivationfunctions}. The smooth gradient allows efficient gradient flow, which is beneficial for deep architectures. Unlike ReLU, swish does not abruptly truncate negative values.

Parametric activation functions emerged as a way to enhance a network's ability to capture complex relationships and outperform fixed, deterministic activation functions.
Parametric ReLU, perhaps the most popular parametric activation function, is the same as LReLU (Eq. \ref{eq:lrelu}), but differs in that $\alpha$ is learned through backpropagation. Swish initially originated as a parametric activation function defined by $f(x)=x\cdot \sigma(\beta x)$ where, on average, it performed the same as the non-parameterized version \citep{ramachandran2017searchingactivationfunctions}. The exponential linear unit, ELU, is also a parametric activation function defined as the identity function for $x>0$ and $\alpha(\exp(x)-1)$ otherwise where the learnable parameter $\alpha$ controls the saturation of ELU for negative inputs thus diminishing the vanishing gradient effect \citep{clevert2016fastaccuratedeepnetwork}. The scaled exponential linear unit, $\text{SELU}(x)=\lambda \cdot \text{ELU}(x)$ builds upon ELU by introducing $\lambda$ and fixing both it and $\alpha$ to carefully derived constants $\alpha \approx 1.67326$ and $\lambda \approx 1.0507$ which has shown to perform exceptionally when the neural network is deep or the number of data points is reasonable ($>1000$) \citep{klambauer2017selfnormalizingneuralnetworks}.

Convolutional neural networks (CNNs) represent a specialized class of artificial neural networks (ANNs) designed to process and analyze visual data. They have achieved 
success in computer vision tasks, which include image recognition, object detection, and semantic segmentation. CNNs leverage a series of convolutional layers to extract spatial hierarchies
of features from input images. The architecture of CNNs typically consists of convolutional layers, pooling layers, and fully connected layers.
For complex deep learning tasks, deeper neural networks were necessary, but research has shown that as the number of layers grows, the harder it is for classical convolutional networks to converge \citep{pmlr-v9-glorot10a}.
This has been solved by CNN-based architectures such as ResNet, DenseNet, and MobileNet by improving gradient flow, reusing features, and optimizing computational efficiency, enabling stable training and scalability for complex tasks.\citep{DBLP:journals/corr/HeZRS15,DBLP:journals/corr/HuangLW16a,DBLP:journals/corr/HowardZCKWWAA17}.

\section{Methodology}

\subsection{Theoretical Foundations}

Eq. \ref{eq:reca} can be rewritten as a piecewise function:

\[f(x)=\begin{cases}
	\alpha x \left(\left(\frac{1+\tanh(x)}{2}\right)^\beta + \sigma(x)^\delta\right) & x>0\\
	0 & \text{otherwise}
\end{cases}\]

\noindent Studying the limits of $\tanh$ and $\sigma$, we observe that as $x \to \infty$, their derivatives approach zero. Consequently, for large $x$, their contributions to the activation function become negligible. The gradient of the $\tanh$ term decays exponentially as $e^{-2x}$, which is twice as fast as the $\sigma$ term, whose gradient decays as $e^{-x}$. Therefore, the $\tanh$ term can be interpreted as a mechanism to fine-tune the function shape for smaller $x$, while the $\sigma$ term provides a broader curvature adjustment.

Next, we seek to understand the smoothness and non-linearity of the function. Due to the trainable parameters $\beta$ and $\delta$, we may control the degree of non-linearity. For zero values of these parameters, ReCA is exactly ReLU (assuming $\alpha = 0.5$), for increasing values, ReCA begins to appear as a non-linear ReLU for $x>0$. To demonstrate this, we generate a grid of 1000 sample points and plot them on a neural network with randomly initialized weights in the range $(-0.05, 0.05)$. To show the difference between ReLU and ReCA, we set the parameters $\beta=\delta=1$ and keep $\alpha=0.5$.

\begin{figure}[!h]
	\centering
	\includegraphics[width=1\linewidth]{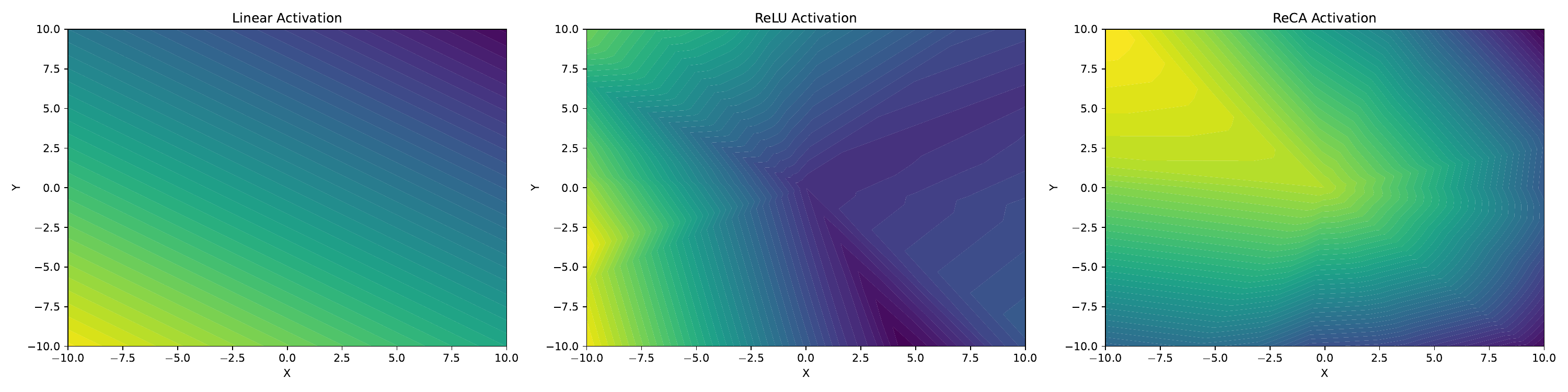}
	\caption{Comparison of Activation Functions' Output Landscapes: The Linear activation (left) displays uniform, globally smooth gradients, while ReLU (middle) introduces sharp, piecewise-linear transitions. ReCA (right) produces smoother and more adaptive gradients, combining nonlinearity with continuity for a richer feature representation.}
	\label{fig:outputs}
\end{figure}

In Figure \ref{fig:outputs}, we demonstrate the output landscape of the linear function, ReLU, and ReCA. The output landscape for the Linear Activation (left panel) shows smooth, uniform gradients across the entire grid, as expected from its globally linear behavior. There are no sharp transitions, and the contours form parallel diagonal lines. This simplicity highlights the lack of nonlinearity, making it unsuitable for capturing complex relationships. The ReLU Activation (middle panel), in contrast, exhibits distinct, sharp transitions with piecewise-linear behavior. The output changes abruptly at certain regions, reflecting the function's zero-gradient inactivation for negative inputs and linear growth for positive inputs. The contour lines are broken and angular, indicative of ReLU’s sparse and piecewise nature. Finally, the ReCA Activation (right panel) demonstrates a smoother, more continuous gradient landscape, combining gradual nonlinear transitions with clear directionality. The contours flow smoothly across the grid, without the abrupt sharpness seen in ReLU. This indicates that ReCA provides a richer, more adaptable feature space while maintaining nonlinearity.

\subsection{Experimental Design}

Our experimental methodology centers on the comprehensive evaluation of the ReCA activation function, as defined in Equation \ref{eq:reca}. The primary objective of these experiments is to assess ReCA's performance across a range of state-of-the-art deep learning models and datasets, in comparison to established activation functions such as ReLU, PReLU, and swish. To initialize the function, we set the scaling parameter $\alpha$ to 0.5 and the smoothing parameters $\beta$ and $\delta$ to 0.05. This choice ensures that ReCA starts as a slightly non-linear version of ReLU, allowing the network to leverage its enhanced flexibility while maintaining familiarity with traditional activation behavior. These parameters are then allowed to adjust dynamically during training through backpropagation, enabling the network to learn the optimal configuration for the given task.

To prevent uncontrolled growth of the trainable parameters $\alpha$, $\beta$, and $\delta$, we apply an L2 regularization penalty with a strength of $10^{-7}$. This regularization term promotes stability during training by discouraging excessively large parameter values that might lead to overfitting or numerical instability. The regularization strength was chosen empirically to balance performance gains with model complexity.

ReCA is designed to function as a drop-in channel-wise replacement for traditional activation functions, making it highly versatile and compatible with existing neural network architectures. This channel-wise implementation allows for a moderate increase in model complexity without significantly impacting memory usage. Importantly, ReCA does not require any extensive modifications to the network architecture, optimizer, or training procedures, making it easy to integrate into various models with minimal effort.
\subsubsection{Datasets}

Image processing represents one of the most challenging and complex domains in deep learning, making it ideal for testing activation functions. We test ReCA on 3 classic datasets: CIFAR-10, CIFAR-100, and Tiny ImageNet, all of which are image-based classification datasets. CIFAR-10 contains 60,000 RGB images of 32x32 size, providing 6,000 images per class
and serving as a great initial benchmark with moderate complexity \citep{c10}. Next, we test on CIFAR-100, maintaining the same total image count but having 100 classes, creating a more challenging classification task with fewer 
images to train on \citep{c100}. Finally, we test ReCA on Tiny ImageNet \citep{Le2015TinyIV}, containing 100,000 images across 200 classes at 64x64 resolution, presenting a significantly more complex task with higher
intra-class variation, which offers a more realistic challenge. Tiny ImageNet is chosen because it provides a challenging yet manageable dataset, balancing complexity and computational feasibility while maintaining a realistic representation of image classification tasks with high intra-class variation and lower resolution compared to the full ImageNet dataset.

\begin{table}[!ht]
	\centering
	\begin{tabular}{lccc}
		\toprule
		Dataset			& Instances		& Size		& Number of Classes \\ \midrule
		CIFAR-10		& 60,000		& 32x32		& 10				\\
		CIFAR-100		& 60,000		& 32x32		& 100				\\
		Tiny ImageNet	& 100,000		& 64x64		& 200				\\ \bottomrule
	\end{tabular}
	\caption{Summary of the Datasets Used}
\end{table}

We split each of these datasets into 80/20 train-test sets and run the same experiment three times on different seeds. This provides sufficient training data to learn meaningful representations, all the while reserving enough testing samples
to ensure reliable evaluation. 

\subsubsection{Model Architectures}

We test ReCA on various state-of-the-art models (shown in more detail in Table \ref{table:configs}).
The ResNet family of models addresses the fundamental challenge of training very deep neural networks by introducing
residual connections that enable better gradient flow. Each variant uses a similar structure of convolutional layers organized in residual blocks. We also train on a Wide ResNet (WRN) which has been shown to outperform very deep resnets while being shallower but wider \citep{wide}.
DenseNet provided an innovative approach for gradient flow within CNNs
by grouping layers into blocks and connecting all layers within a block to each other, allowing each layer to receive feature maps from all preceding layers, which significantly enhances feature reuse and gradient flow.
In our experiments, we use the DenseNet-BC-121 model, where the BC designation refers to the bottleneck modification, which reduces the number of feature maps before expensive convolution operations which forces the network to learn more efficient representations
of the network data. Finally, we test ReCA on MobileNetV3, an efficient neural network architecture specifically designed for mobile and embedded devices.

In our experiments, we train all these models using cosine annealing \citep{anneal} with lower bound $10^{-4}$ to ensure smooth learning and gradient flow.

\begin{table}[!ht]
	\centering
	\begin{tabular}{lcccc}
		\toprule
		Model			& Epochs		& Batch Size		& Optimizer		& Learning Rate \\ \midrule
		\midrule
		ResNet-20		& 200			& 128				& SGD			& 0.05 \\
		WRN-16-8		& 200			& 128				& SGD			& 0.05 \\
		ResNet-32		& 200			& 128				& SGD			& 0.05 \\
		ResNet-56		& 200			& 128				& SGD			& 0.05 \\
		DenseNet-BC-121 & 250			& 64				& Adam			& 0.001 \\
		MobileNetV3		& 250			& 64				& SGD			& 0.05 \\
		\bottomrule
	\end{tabular}
	
	\caption{Model Configurations}
    \label{table:configs}
\end{table}

\pagebreak

\section{Results}

Across a variety of datasets and architectures, ReCA consistently demonstrates superior performance compared to traditional and parametric activation functions, which shows its robustness across different tasks and architectures. ReCA is tested against ReLU, PReLU, and swish. ReCA's ability to adapt to the given task and architecture results in improved feature representation and higher accuracy.

\subsection{CIFAR}

\begin{figure}[!h]
	\centering
	\includegraphics[width=1\linewidth]{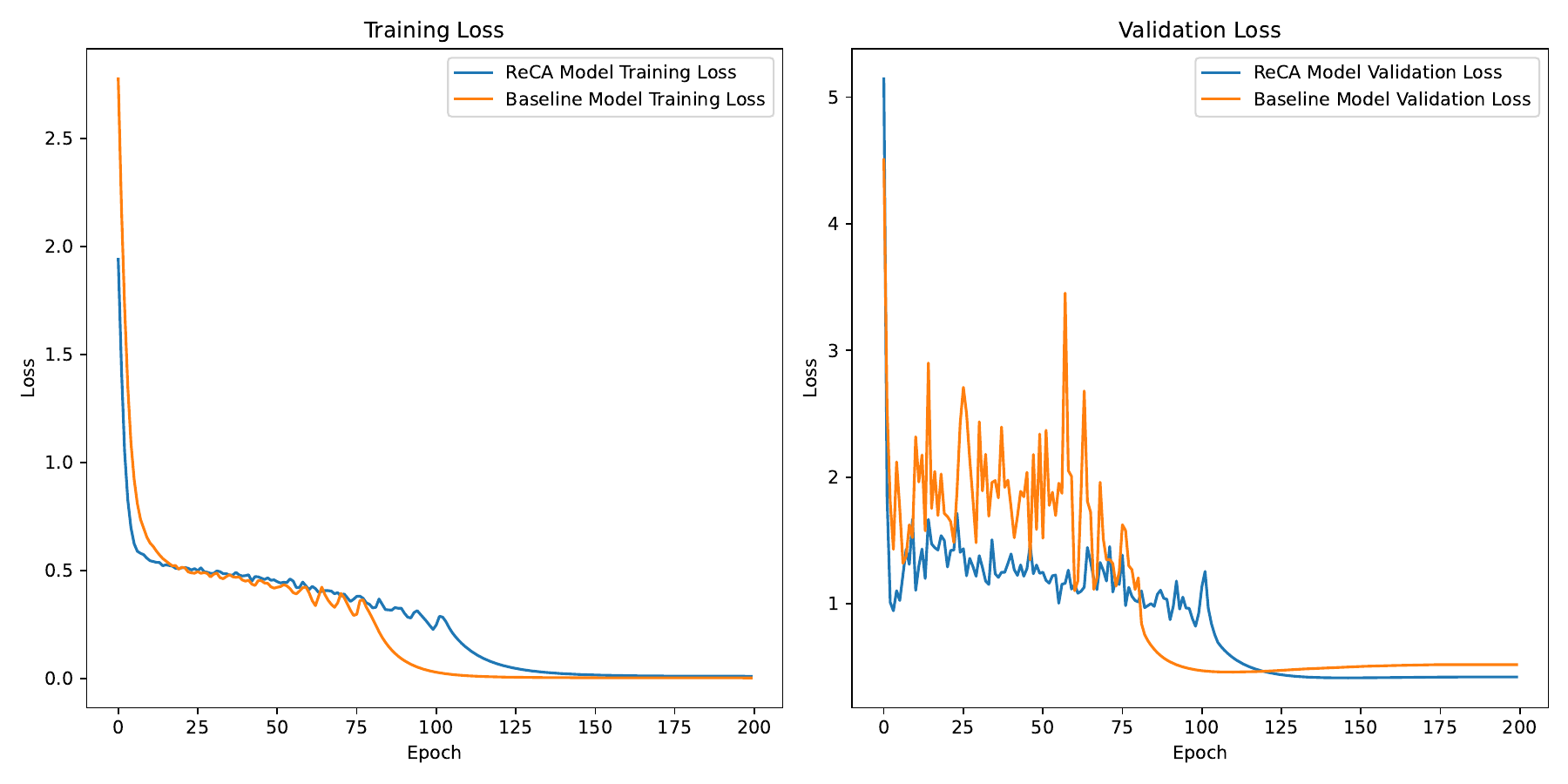}
	\caption{Train and validation loss of PReLU (baseline) vs ReCA on CIFAR-100}
	\label{fig:preluvsreca}
\end{figure}

The CIFAR-10 dataset, with its 10 classes and 32x32 images, serves as a foundation for assessing the performance of ReCA on a relatively simple classification task. Table \ref{tab:cifar10resnet20} presents the top-1 accuracy of the ResNet-20 model, comparing ReLU, PReLU (applied per neuron), swish, and ReCA (applied per channel). Across the three runs, ReLU has an average accuracy of 83.78\%, PReLU of 82.97\%, and swish of 82.10\%. ReCA outperforms all the baselines, showing an average accuracy of 85.90\%. Similarly, on the Wide ResNet 16-8 model, ReCA outperforms all baselines, showing a 1.24\% accuracy jump over the next best activation function: ReLU. While PReLU under-performing ReCA may be seen as an overfitting issue since the PReLU model is over-parameterized, Figure \ref{fig:preluvsreca} shows that even with significantly fewer parameters, ReCA can achieve the same train loss as PReLU while retaining generalization capabilities.

CIFAR-100 proves to be a significantly harder classification task than CIFAR-10, containing just 600 samples per class. In this experiment, ReCA significantly outperforms all baselines, showing a 4.59\% average increase in performance over PReLU (best-performing baseline) when tested on ResNet-32. On ResNet-56, the best-performing baseline is swish with an average accuracy of 51.11\%. ReCA outperforms this by 5.19\%, showcasing an average accuracy of 56.30\%.

\begin{figure}[!ht]
	\centering
	\includegraphics[width=1\linewidth]{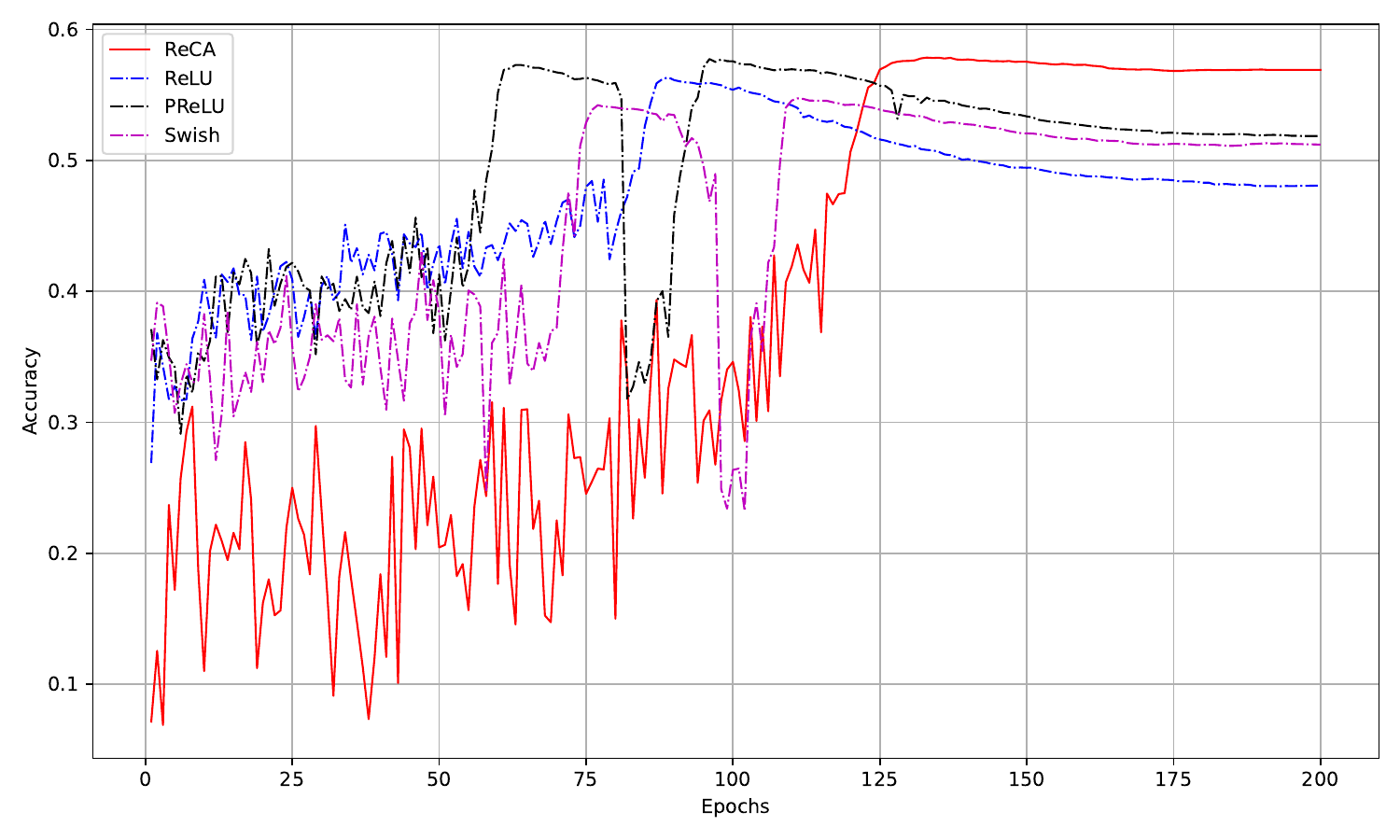}
	\caption{ResNet-56 (median-of-three) performance on CIFAR-100 validation data per epoch}
	\label{fig:validation}
\end{figure}

By design, ReCA outputs zero for all negative inputs and learns an appropriate curvature for all positive inputs through backpropagation. These two factors encourage sparsity and allow smooth gradient flow, respectively, explaining why ReCA significantly outperforms all baselines. Table \ref{tab:cifar10wrn168} further backs up this hypothesis, as the only difference between ReLU (top-performing baseline) and ReCA (top-performing function) is the learned non-linearity for positive inputs. Figure \ref{fig:validation} shows that while ReCA requires more epochs to reach peak accuracy, its performance does not degrade at all afterward, unlike the baseline functions.

\begin{table}[htbp]
	\centering
	\begin{minipage}{0.45\linewidth}
		\centering
		\begin{tabular}{lccc}
			\toprule
			Function & Run 1 & Run 2 & Run 3 \\ \midrule
			ReLU     & 83.96 & 84.11 & 83.27 \\
			PReLU (per-neuron)    & 83.02 & 83.39 & 82.52 \\
			Swish    & 82.24 & 81.85 & 82.22 \\
			ReCA (channel-wise)    & \textbf{85.07} & \textbf{86.21} & \textbf{86.41} \\ \bottomrule
		\end{tabular}
		\vspace{0.1in}
		\caption{Top-1 accuracy of ResNet-20 on CIFAR-10}
		\label{tab:cifar10resnet20}
	\end{minipage}%
	\hfill
	\begin{minipage}{0.45\linewidth}
		\centering
		\begin{tabular}{lccc}
			\toprule
			Function & Run 1 & Run 2 & Run 3 \\ \midrule
			ReLU     & 88.69 & 86.83 & 86.71 \\
			PReLU (per-neuron)    & 87.30 & 86.04 & 86.00 \\
			Swish    & 88.49 & 88.48 & 81.56 \\
			ReCA (channel-wise)    & \textbf{88.70} & \textbf{88.52} & \textbf{88.73} \\ \bottomrule
		\end{tabular}
		\vspace{0.1in}
		\caption{Top-1 accuracy of WRN16-8 on CIFAR-10}
		\label{tab:cifar10wrn168}
	\end{minipage}
\end{table}

\begin{table}[!ht]
	\centering
	\begin{minipage}{0.45\linewidth}
		\centering
		\begin{tabular}{lccc}
			\toprule
			Function & Run 1 & Run 2 & Run 3 \\ \midrule
			ReLU     & 47.90 & 51.29 & 51.34 \\
			PReLU (per-neuron)   & 46.68 & 52.00 & 53.49 \\
			Swish    & 45.84 & 45.57 & 46.47 \\
			ReCA (channel-wise)    & \textbf{53.65} & \textbf{55.54} & \textbf{56.76} \\ \bottomrule
		\end{tabular}
		\vspace{0.1in}
		\caption{Top-1 accuracy of ResNet-32 on CIFAR-100}
		\label{tab:cifar100resnet32}
	\end{minipage}%
	\hfill
	\begin{minipage}{0.45\linewidth}
		\centering
		\begin{tabular}{lccc}
			\toprule
			Function & Run 1 & Run 2 & Run 3 \\ \midrule
			ReLU     & 47.17 & 48.05 & 51.99 \\
			PReLU (per-neuron)    & 46.81 & 54.68 & 51.84 \\
			Swish    & 49.02 & 51.20 & 55.09 \\
			ReCA (channel-wise)    & \textbf{54.70} & \textbf{56.90} & \textbf{57.31} \\ \bottomrule
		\end{tabular}
		\vspace{0.1in}
		\caption{Top-1 accuracy of ResNet-56 on CIFAR-100}
		\label{tab:cifar100resnet56}
	\end{minipage}
\end{table}

\pagebreak

\subsection{Tiny ImageNet}

Tiny ImageNet presents a significantly more complex challenge than the two previously encountered. We use the DenseNet-BC-121 model, known for its efficient feature reuse, to evaluate ReCA's performance as shown in Table \ref{tab:tinyimagenetdensenet}. ReCA consistently outperforms ReLU in both top-1 and top-5 accuracy. Top-1 accuracy for ReCA ranged from 41.29\% to 41.80\%, while that of ReLU ranged from 39.95\% to 40.74\%. The top-5 accuracies show a similar trend, with ReCA consistently outperforming ReLU.

MobileNetV3-Small, designed for efficiency, was also used to evaluate ReCA's performance. Here, ReCA again consistently outperforms ReLU in both top-1 and top-5 accuracy as shown in Table \ref{tab:tinyimagenetmobilenet}. This indicates that even in efficient models designed for mobile devices, ReCA provides a noticeable improvement in accuracy.

The hypothesis proposed in the preceding section is further verified as the non-linearity of ReCA for positive inputs allows for greater flexibility and representational capability.

\begin{table}[!ht]
	\centering
	\begin{tabular}{lcccccc}
		\toprule
		& \multicolumn{2}{c}{Run 1} & \multicolumn{2}{c}{Run 2} & \multicolumn{2}{c}{Run 3} \\ 
		\cmidrule(lr){2-3} \cmidrule(lr){4-5} \cmidrule(lr){6-7}
		Function & Top-1 & Top-5 & Top-1 & Top-5 & Top-1 & Top-5 \\ \midrule
		ReLU        & 39.95 & 63.9 & 40.43 & 64.42 & 40.74 & 64.35 \\
		ReCA (channel-wise)          & \textbf{41.29}  & \textbf{65.06}  & \textbf{41.56}  & \textbf{65.21}  & \textbf{41.80}  & \textbf{65.38}  \\
		\bottomrule
	\end{tabular}
	\vspace{0.1in}
	\caption{Top-1 and Top-5 accuracy of DenseNet-BC-121 on Tiny ImageNet}
	\label{tab:tinyimagenetdensenet}
\end{table}

\begin{table}[!ht]
	\centering
	\begin{tabular}{lcccccc}
		\toprule
		& \multicolumn{2}{c}{Run 1} & \multicolumn{2}{c}{Run 2} & \multicolumn{2}{c}{Run 3} \\ 
		\cmidrule(lr){2-3} \cmidrule(lr){4-5} \cmidrule(lr){6-7}
		Function & Top-1 & Top-5 & Top-1 & Top-5 & Top-1 & Top-5 \\ \midrule
		ReLU        & 25.67 & 48.42 & 24.48 & 47.69 & 23.74 & 46.77 \\
		ReCA (channel-wise)          & \textbf{25.72}  & \textbf{49.50}  & \textbf{25.51}  & \textbf{48.64}  & \textbf{24.86}  & \textbf{47.77}  \\
		\bottomrule
	\end{tabular}
	\vspace{0.1in}
	\caption{Top-1 and Top-5 accuracy of MobileNetV3-Small on Tiny ImageNet}
	\label{tab:tinyimagenetmobilenet}
\end{table}

Table \ref{table:timeandspace} compares the resource requirements for ReLU and ReCA. While ReCA introduces a negligible average increase in parameters (0.84\%), the computational overhead during training is more pronounced. We observe an average increase in training time of 52.20\%. However, this figure is significantly skewed by the DenseNet-BC-121 experiment, which exhibited a disproportionately large increase of over 2.5x. For other architectures, the training time increase is more moderate, typically ranging from 17\% to 40\%. This additional training cost represents a direct trade-off for the performance improvements and more stable gradient flow offered by ReCA,
which is particularly valuable in achieving high accuracy in deep models.

\begin{table}[htbp]
	\centering
	\begin{tabular}{lcc|cc}
		\toprule
		Dataset & \multicolumn{2}{c|}{ReLU} & \multicolumn{2}{c}{ReCA} \\
		\cmidrule(rr){2-3} \cmidrule(rr){4-5}
				& Params & Time & Params & Time \\ \midrule
				\textbf{CIFAR-10} \\
		\hspace{1em} ResNet-20 & 540,882 & 9.07 min & 544,950 & 12.38 min \\
		\hspace{1em} WRN16-8   & 21,929,942 & 83.99 min & 21,951,542 & 98.55 min \\
		\textbf{CIFAR-100} \\
		\hspace{1em} ResNet-32 & 942,282 & 13.81 min & 949,098 & 19.33 min \\
		\hspace{1em} ResNet-56 & 1,721,802 & 23.01 min & 1,733,994 & 31.81 min \\
		\textbf{Tiny ImageNet} \\
		\hspace{1em} DenseNet-BC-121 & 30,224,538 & 4.45h & 30,534,138 & 11.54h \\
		\hspace{1em} MobileNetV3-Small & 802,856 & 1.91h & 817,220 & 2.67h \\
		\bottomrule
	\end{tabular}
	\vspace{0.1in}
	\caption{Comparison of Space and Time taken for ReLU and ReCA across datasets and models.}
	\label{table:timeandspace}
\end{table}

\section{Conclusion}

In this paper, we introduced ReCA, a novel parametric activation function designed to combine the strengths of classical activation functions like ReLU, $\tanh$, and sigmoid while addressing their limitations. Extensive experiments across diverse datasets and architectures demonstrated that ReCA consistently outperforms traditional activation functions, achieving higher accuracy and providing smoother gradients for deeper architectures. 

Despite these benefits, ReCA incurs additional computational overhead, with a modest increase in model parameters and training time compared to ReLU. Future work will focus on optimizing ReCA's computational efficiency and exploring its applicability to other tasks beyond image classification. ReCA's ability to enhance performance while maintaining flexibility highlights its potential as a valuable tool in the development of advanced deep learning models.

\bibliography{chidiac-azar-reca-fixed}

\end{document}